\documentclass[10pt,twocolumn,letterpaper]{article}

\usepackage{subfigure}
\usepackage{amsmath}
\usepackage{cvpr}
\usepackage{times}
\usepackage{epsfig}
\usepackage{graphicx}
\usepackage{amsmath}
\usepackage{amssymb}
\usepackage{booktabs}

\usepackage[pagebackref=true,breaklinks=true,letterpaper=true,colorlinks,bookmarks=false]{hyperref}

\cvprfinalcopy 

\def\cvprPaperID{4096} 

\ifcvprfinal\fi
\begin{document}
\title{Exploring Hypergraph Representation on Face Anti-spoofing Beyond 2D Attacks}

\author{Wei Hu\\
Peking University\\
{\tt\small forhuwei@pku.edu.cn}
\and
Gusi Te\\
Peking University\\
{\tt\small tegusi@pku.edu.cn}
\and
Ju He\\
Peking University\\
{\tt\small heju@pku.edu.cn}
\and
Dong Chen\\
Microsoft Research Asia\\
{\tt\small doch@microsoft.com}
\and
Zongming Guo\\
Peking University\\
{\tt\small guozongming@pku.edu.cn}
}

\maketitle

\begin{abstract}
  Face anti-spoofing plays a crucial role in protecting face recognition systems from various attacks. Previous model-based and deep learning approaches achieve satisfactory performance for 2D face spoofs, but remain limited for more advanced 3D attacks such as vivid masks. In this paper, we address 3D face anti-spoofing via the proposed Hypergraph Convolutional Neural Networks (HGCNN). Firstly,  we construct a computation-efficient and posture-invariant face representation with only a few key points on hypergraphs. The hypergraph representation is then fed into the designed HGCNN with hypergraph convolution for feature extraction, while the depth auxiliary is also exploited for 3D mask anti-spoofing. Further, we build a 3D face attack database with color, depth and infrared light information to overcome the deficiency of 3D face anti-spoofing data. Experiments show that our method achieves the state-of-the-art performance over widely used 3D and 2D databases as well as the proposed one under various tests.          
\end{abstract}

\section{Introduction}
\begin{figure}[htbp]
    \centering
    \includegraphics[width=0.4\textwidth]{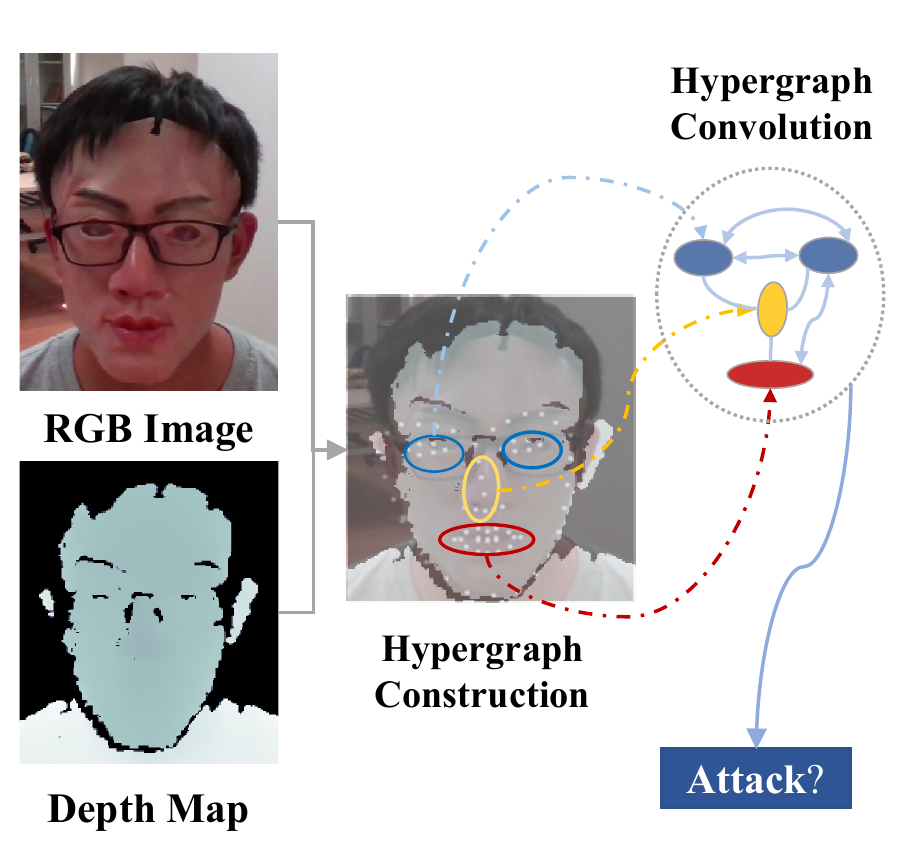}
    \vspace{-0.05in}
    \caption{Illustration of the proposed HGCNN architecture for 3D face anti-spoofing, which takes RGB and depth images as the input and generates classification scores as the output. We first construct a hypergraph over augmented landmarks of each face on both the RGB and depth images. Then we perform the designed hypergraph convolution on both modalities for feature extraction, which leads to the final spoofing scores. Our model is posture-invariant and computation-efficient due to the robustness and compactness of the proposed hypergraph representation. (Best viewed in color)}
    \label{fig:title}
\end{figure}

Face recognition has been widely applied to a variety of areas, including access control systems, online payment and user authentication. Nevertheless, vulnerability exists in a large amount of systems that they sometimes fail to recognize fake faces, which may be used by attackers to hack the systems. This is also referred to as \textit{face spoofing}, an attempt to deceive recognition systems with photos, videos or masks. In this paper, we aim to address the following types of attacks, with emphasis on 3D masks:

\textbf{Print Attack-2D}: An attacker uses a photo printed or displayed on the screen to deceive the camera. This attack only employs one fixed image.

\textbf{Replay Attack-2D}: This attack is more tricky with a period of video played repeatedly in front of the camera. Human behavior is more natural compared to print attacks.

\textbf{Mask Attack-3D}: In this type of attacks, a mask is utilized to conceal the original face. The success of attacks depends on the quality of masks. The state-of-the-art masks\footnote{For example, ThatsMyFace.com makes unique personalized lifesize wearable masks with one's likeness from photos.} are so vivid that sometimes even human may be unable to discriminate real faces. 

Various methods have been proposed to address these attacks, i.e., \textit{face anti-spoofing}. Previous model-based approaches \cite{Määttä11ijcb,Pereira12accv,Pereira13icb,yang13icb,RN41,Patel16tifs,Boulkenafet17spl,Mirjalili17icb} make use of RGB images or sequences as the input and feed hand-crafted features such as LBP features into classifiers, which lacks 3D information. With the development of deep learning, Convolutional Neural Network (CNN) becomes a powerful tool for feature extraction, and is thus leveraged for face anti-spoofing \cite{feng16jvcir,li16ipta,patel2016cross,Liu18CVPR}. Several approaches have achieved promising performance via CNN and other techniques such as optical flow analysis \cite{RN43}. Depth is a kind of remarkable cue for recognizing 2D attacks such as print and replay attacks. Atoum et al. propose to estimate the depth of a face image via a fully convolutional network, which is then utilized for anti-spoofing \cite{RN29}. Instead of depth estimation, Wang et al. capture depth maps from Kinect for classification along with 2D facial images \cite{WANG2017332}. However, both methods leverage depth for detecting 2D attacks, while 3D mask attacks remain to be addressed.    

Hence, we propose Hypergraph Convolution Neural Networks (HGCNN) with RGB-D information to detect both 2D attacks (print/replay) and 3D mask attacks. Firstly, we propose compact face representation by only key points, including landmarks and a few interpolated points, which leads to remarkable reduction of both time and space complexity while retaining most facial information. Because the key points are likely to reside on irregular grids, we further represent them on a hypergraph, which models \textit{high-order relationship} of samples via hyperedges, each of which links multiple samples. As the relationship among samples on a hypergraph is \textit{invariant to postures}, i.e., the relationship remains the same no matter how one moves the face, hypergraph-based facial representation is more robust to face motions or emotions compared with traditional image representation. This is because the hypergraph describes the relative relationship instead of the absolute Euclidean distance among samples.  

Thereafter, we take the hypergraph representation of faces as the input, and design a framework of HGCNN based on hypergraph convolution. Essentially this is an extension to the recently proposed Graph Convolutional Neural Network (GCNN) \cite{bruna2013spectral,duvenaud2015convolutional,kipf2016semi,defferrard2016convolutional}, which takes both data features and connectivity/topology\footnote{The connectivity/topology of graph signals is often represented by an adjacency matrix or a graph Laplacian matrix. The definitions of the matrices in spectral graph theory will be introduced in Section~\ref{sec:pre}.} of graph signals as the input and extracts higher-level features via graph convolution, either in spectral domain or in nodal domain. However, the input graphs in the current GCNNs are simple graphs, where an edge connects just two samples for modeling the pairwise relationship. Also, simple graphs are sensitive to the radius parameter used in the similarity calculation. Hence, we extend the input simple graphs to hypergraphs, which address both problems by modeling higher-order relationship among samples via hyperedges. The subsequent graph convolution, is thus also extended to hypergraph convolution for higher-order feature extraction. Further, we exploit the depth auxiliary for 3D mask anti-spoofing, where depth maps share the same hypergraph representation as the RGB cue and then go through hypergraph convolution for feature learning. The extracted depth features are then concatenated with the textural features to form the final node features for classification.     

Also, we build a 3D face attack database containing color, depth and Infrared ray (IR) data acquired from Intel RealSense SR300\footnote{https://realsense.intel.com/}, which embodies more subjects and variations than prior 3D face databases. Extensive experiments show our method has superior performance and robustness on existing widely used 2D/3D databases and the proposed one. 

In summary, our contributions mainly include: 
\begin{itemize}
	\item We propose hypergraph-based face representation with only a few key points, which is computation-efficient and posture-invariant. Based on the hypergraph representation, we design HGCNN with hypergraph convolution for feature extraction.    
    \item We exploit both RGB and depth information for 3D mask anti-spoofing. The depth auxiliary is fused with texture in the feature domain from hypergraph convolution.   
    \item We collect a 3D face attack database of rich and diverse information, and achieve state-of-the-art performance on widely used 2D/3D databases and the proposed one. 
\end{itemize}
    
\section{Related Work}
\subsection{Face anti-spoofing}
Previous face anti-spoofing methods can be classified into spatial methods, temporal methods and fusion methods. 

\textbf{Spatial methods.} Texture is a good hint for discriminating between real faces and fake ones, since print or replay attacks exhibit different textural characteristics from real faces. Li et al. is the first to take frequency distribution into consideration \cite{RN36}. Other hand-crafted features are introduced to tackle face anti-spoofing, such as LBP \cite{RN36,RN37,ERDOGMUS_BTAS-2013}, HoG \cite{RN41} and DoG \cite{RN40}. After extracting high-level features, they adopt classifiers for final results, among which SVM is a typical one \cite{RN37}. Moreover, Galbally et al. exploit the ability of image quality from the background and face to improve the results \cite{RN42}. 

With the rapid development of deep learning, CNN-based anti-spoofing methods have been proposed to extract features. Most of them regard the problem as simple binary classification and extract features from texture images by traditional networks such as VGG or ResNet \cite{li16ipta,patel2016cross}. Furthermore, \cite{RN29} adopts depth as auxiliary information, divides images into different patches and feeds depth and color cues into the network. 

\textbf{Temporal methods.} Several methods explore the potential of liveness detection from temporal sequences, including head movements and facial expressions. Pan et al. propose a straight-forward method that utilizes eye-blinking to detect whether the facial motion is authentic\cite{RN39}. Besides, optical flow is introduced to analyze tiny expressions, which is essential to extract rigid movements of masks \cite{RN43}. Edmunds et al. extract low-level motion features such as eye gazing and head pose from video clips to integrate high-level features \cite{RN28}. Xu et al. \cite{Xu2015LearningTF} propose an LSTM + CNN framework to get fusion features for anti-spoofing.

\textbf{Fusion methods.} Leveraging on existing methods, some approaches combine texture and temporal cues, which achieve significant performance. Asim et al. propose a CNN-based spatial-temporal feature extraction framework for classification\cite{RN46}. In \cite{RN29}, Atoum et al. propose a multi-cue integration framework, including image quality, optical flow map and LBP features, followed by a neural network classifying integrated features. Furthermore, Liu et al. exploit Remote photoplethysmography (rPPG), a kind of signal reflecting facial liveness, and propose a neural network combining CNN and RNN to generate rPPG signals for liveness detection\cite{Liu18CVPR}.

\subsection{Graph Convolutional Neural Networks}
GCNN extends CNN by consuming data defined on irregular grids. The key challenge is to define convolution over graphs, which is difficult due to the unordered data. According to the definitions of graph convolution, GCNN can be classified into spectral-domain methods and nodal-domain method.   

\textbf{Spectral-domain methods.} The convolution over graphs is elegantly defined in the spectral domain, which is the multiplication of the spectral-domain representation of signals. Specifically, the spectral representation is in the graph Fourier transform (GFT) \cite{hammond2011wavelets} domain, where each signal is projected onto the eigenvectors of the graph Laplacian matrix \cite{hammond2011wavelets, henaff2015deep}. The computation complexity, however, is high due to the eigen-decomposition of the graph Laplacian matrix in order to get the eigenvector matrix. Hence, it is improved by \cite{defferrard2016convolutional} through fast localized convolutions, where the Chebyshev expansion is deployed to approximate GFT. Besides, Susnjara et al. introduce the Lancoz method for approximation \cite{susnjara2015accelerated}. Spectral GCNN has shown its efficiency in various applications such as segmentation and classification \cite{kipf2016semi, Te18mm}.
    
\textbf{Nodal-domain methods.} Many techniques are introduced to implement graph convolution directly on each node and its neighbors. Gori et al. introduce recurrent neural networks that operate on graphs in \cite{gori2005new}. Duvenaud et al. propose a convolution-like propagation to accumulate local features \cite{duvenaud2015convolutional}. Bruna et al. deploy the multi-scale clustering of graphs in convolution to implement multi-scale representation \cite{bruna2013spectral}. Furthermore, Niepert et al. define convolution on a sequence of nodes and perform normalization afterwards \cite{niepert2016learning}. Wang et al. propose edge convolution on graphs by incorporating local neighborhood information \cite{wang2018dynamic}, which is applied to point cloud segmentation and classification. Nodal-domain methods provide strong localized filters, which however also means it is difficult to learn the global structure.

\section{Preliminaries}
\label{sec:pre}
A hypergraph $\mathcal{G}=\{\mathcal{V},\mathcal{E},\mathbf{W}\}$ consists of a vertex set $\mathcal{V}$, a hyperedge set $\mathcal{E}$ where each hyperedge $e_i$ is assigned a weight $w(e_i)$, and a diagonal matrix of the hyperedge weights $\mathbf{W}$. Further, $\mathcal{G}$ can be represented by a $|\mathcal{V}| \times |\mathcal{E}|$ matrix $\mathbf{H}$, with entries $ h(v,e)=1 $ if $v \in e$ and $0$ otherwise, which is referred to as the incidence matrix of $\mathcal{G}$. Based on $\mathbf{H}$, the degree of each vertex $v \in \mathcal{V}$ is 
\begin{equation}
d(v) = \sum_{e \in \mathcal{E}}w(e)\mathbf{H}(v,e),  
\end{equation}
whereas the degree of each hyperedge $e \in \mathcal{E}$ is 
\begin{equation}
\delta(e) = \sum_{v \in \mathcal{V}}\mathbf{H}(v,e).  
\end{equation}
For $k$-uniform hypergraphs considered in our context, the degrees of all the hyperedges are the same, i.e., $\delta(e_i) = k, \forall e_i \in \mathcal{E}$. We then let $\mathbf{D}_v$ and $\mathbf{D}_e$ denote the diagonal matrices containing the vertex and hyperedge degrees, respectively.  

Several definitions of the hypergraph Laplacian have been proposed, including clique expansion \cite{zien99}, star expansion \cite{zien99}, Bolla's Laplacian \cite{Bolla93}, Rodriquez's Laplacian \cite{Rodriguez02,Rodriguez03} and normalized Laplacian \cite{Zhou05}. It has been analyzed in \cite{zhou07} that these formulations are similar to each other. In the context of graph convolutional neural networks, we employ the normalized Laplacian in \cite{RN35} because of its normalization property, which is defined as 
\begin{equation}
\mathcal{L} = \mathbf{I}-\mathbf{D}_v^{-\frac{1}{2}}\mathbf{H}\mathbf{W}\mathbf{D}_e^{-1}\mathbf{H}^T\mathbf{D}_v^{-\frac{1}{2}}.
\label{eq:hyperL}
\end{equation}

\section{Approach}
We first overview the architecture of the proposed HGCNN. Then we dive into our method from the crucial hypergraph representation to the subsequent hypergraph convolution and feature learning. 

\begin{figure*}[t]
	\centering
	\includegraphics[width=0.9\textwidth]{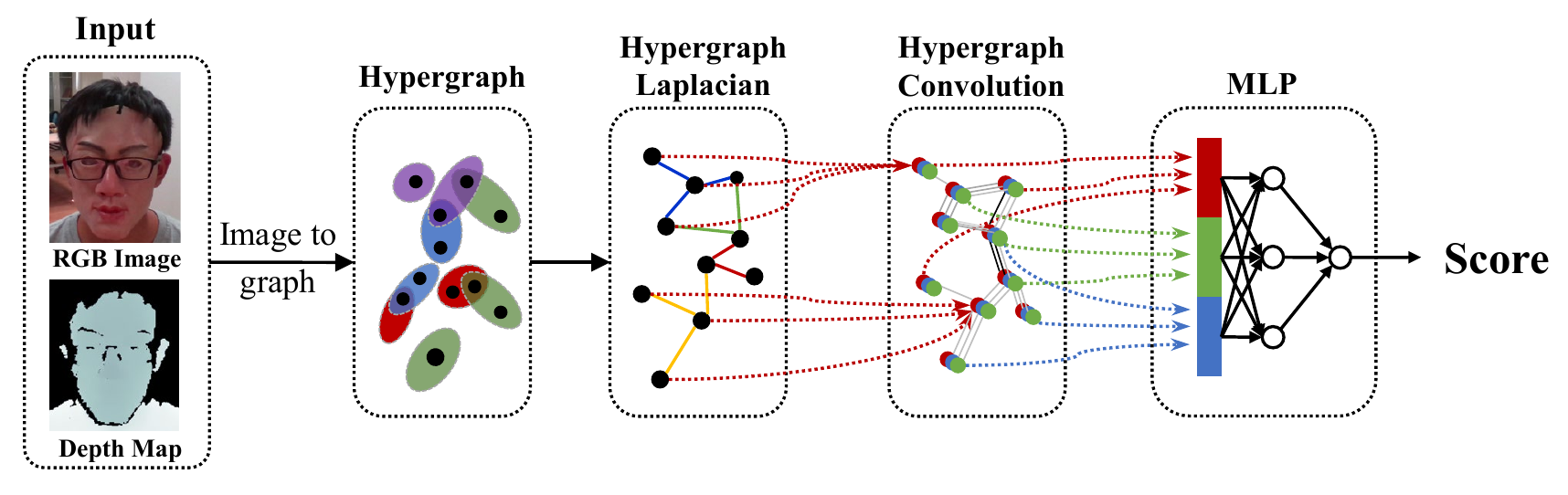}
	\caption{The framework of our proposed HGCNN given a pair of RGB and depth images for 3D face anti-spoofing.}
	\label{fig:pipeline1}
	\vspace{-0.5cm}
\end{figure*}

\subsection{HGCNN Architecture}

\begin{figure}[t]
	\centering
	\includegraphics[width=0.5\textwidth]{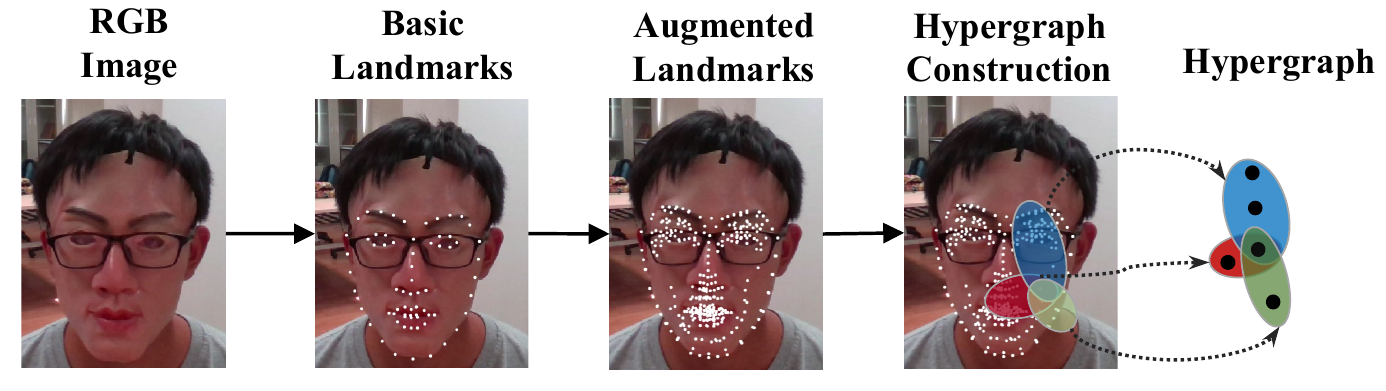}
	\caption{Landmark-based hypergraph construction from a given RGB image.}
	\label{fig:pipeline3}
    \vspace{-0.5cm}
\end{figure}

Fig.~\ref{fig:pipeline1} illustrates the pipeline of the proposed framework. The input consists of RGB and depth images. Firstly, we construct a hypergraph on the input, in which we extract landmarks from the RGB image and then augment the extracted landmarks for the purpose of denser point sets, which conduces to learning local features, as demonstrated in detail in Fig.~\ref{fig:pipeline3}. We then construct a $k$-uniform hypergraph over the landmarks according to Euclidean distance of point pairs, from which we compute the hypergraph Laplacian. Secondly, we feed the RGB and depth features of the landmarks into separate branches of hypergraph convolution, along with the computed hypergraph Laplaican for learning high-level textural and depth features respectively. Finally, we concatenate the extracted features and use Multi-Layer Perceptron (MLP) to acquire the output classification scores.  

\subsection{Hypergraph Representation}

Unlike existing methods, our network takes graphs rather than images as the input, which means the hypergraph structure is the key to the subsequent neural network. In graph-based image representation, a pixel is often treated as a vertex in the graph, and similar pixels are connected via edges. Due to the enormous amount of pixels in an image, it is extremely time-consuming to take every pixel as the input. Instead, inspired by facial landmark detection \cite{RN50}, we propose an efficient approach for hypergraph construction based on limited number of facial landmarks, as illustrated in Fig.~\ref{fig:pipeline1}. 

\textbf{Landmark detection and augmentation.} There are plenty of methods for face landmark extraction. In order to extract landmarks rapidly and robustly, we apply \cite{RN50} to the input RGB image and obtain 68 landmarks. If the landmarks are out of box or not detected, the current frame is neglected. Since the number of landmarks is not enough for feature extraction from neural networks, point augmentation is necessary. We thus propose to augment points by interpolation of the detected landmarks. 

We firstly calculate $k$-nearest-neighbors of each landmark, and add the midpoint of each neighboring pair to the point set as augmentation. The distance metric between a pair of points $\{i,j\}$ is defined as the Euclidean distance, i.e., $ a_{i,j}= \|\mathbf{p}_i-\mathbf{p}_j\|_2^2$, where $\mathbf{p}_i$ and $\mathbf{p}_i$ are the coordinates of point $i$ and $j$ respectively. As some midpoints might be overlapping, we eliminate redundant points, resulting in 250 interpolated points. Together with the original landmarks, we finally extract a total of 318 points with RGB and depth cues to represent each face for the subsequent processing. Note that, as the detected landmarks reside on irregular grids in general, all the extracted points are also irregular, which is difficult to represent via traditional signal representation. 

\textbf{Hypergraph Construction.} In order to describe the high-order relationship of the extracted irregular landmarks, we propose to construct a hypergraph over each face. Specifically, we treat the extracted points as vertices on the hypergraph, and connect each point with its $k$-nearest-neighbors using an hyperedge, which leads to a $(k+1)$-uniform hypergraph. The weight for each hyperedge is assigned $1$, as we assume all the hyperedges are of equal importance.

\subsection{Hypergraph Convolution}
The core of HGCNN is hypergraph convolution, which is an extension to graph convolution. Graph convolution has received much attention recently, and many network models have been proposed to tackle graph-based problems \cite{kipf2016semi}. Unlike images or videos, it is difficult to define convolution over graphs/hypergraphs in the vertex domain, because a meaningful translation operator in the vertex domain is nontrivial to define due to the unordered vertices. Inspired by \cite{defferrard2016convolutional}, we start from filtering of hypergraph signals in the spectral domain, and then deploy Chebyshev approximation to reduce the computational complexity.

\begin{figure*}[t]
\begin{center}
\begin{tabular}{ccccccc}
 \textbf{Blink} & \textbf{Translation} & \textbf{Move back} & \textbf{Expression} & \textbf{Rotation} & \textbf{Replay} & \textbf{Print}\\
\includegraphics[width=0.12\textwidth,height=0.16\textwidth]{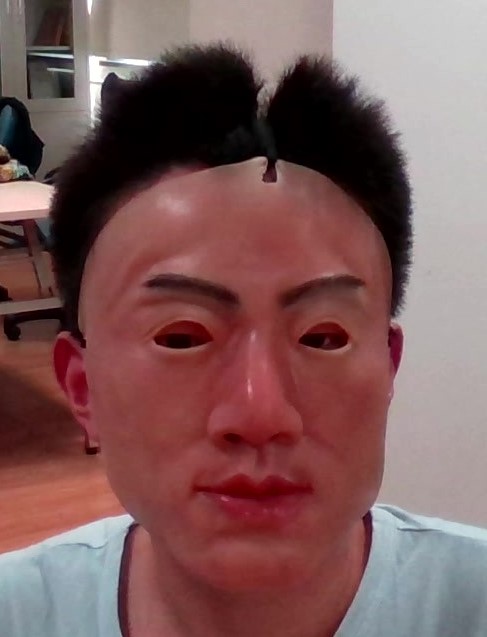} & 	\includegraphics[width=0.12\textwidth,height=0.16\textwidth]{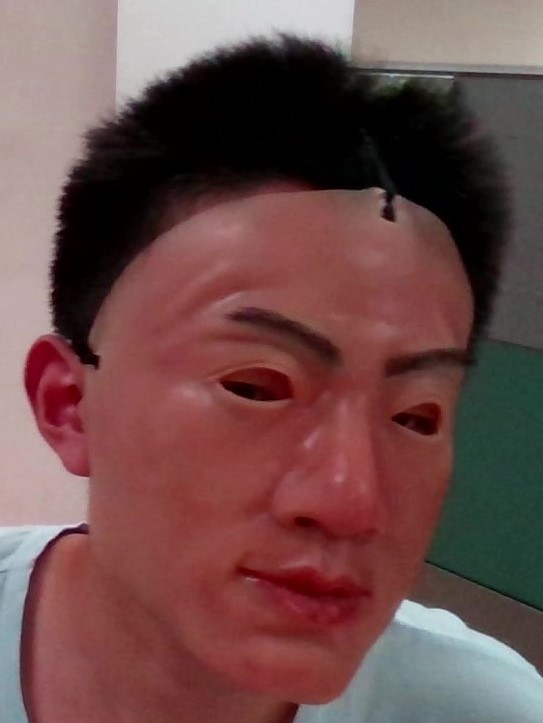}  &  \includegraphics[width=0.12\textwidth,height=0.16\textwidth]{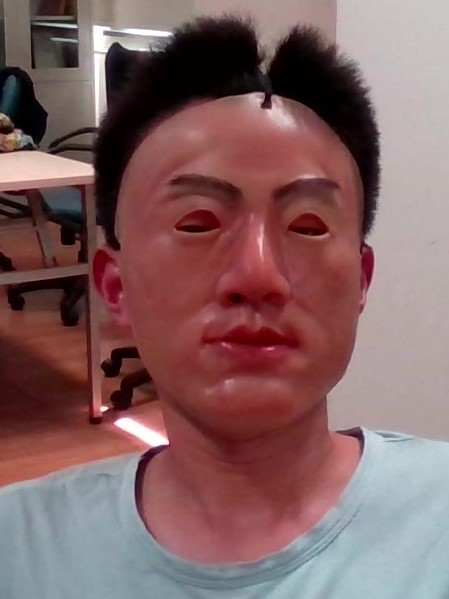} &  \includegraphics[width=0.12\textwidth,height=0.16\textwidth]{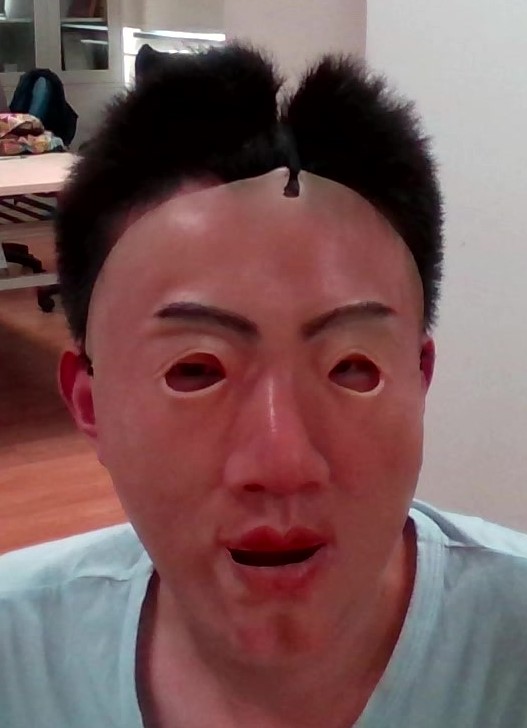}  &  \includegraphics[width=0.12\textwidth,height=0.16\textwidth]{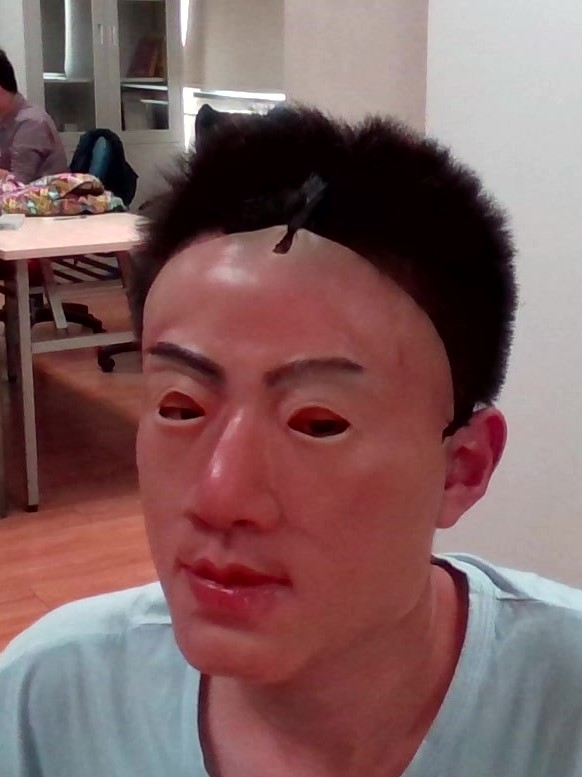} &  \includegraphics[width=0.12\textwidth,height=0.16\textwidth]{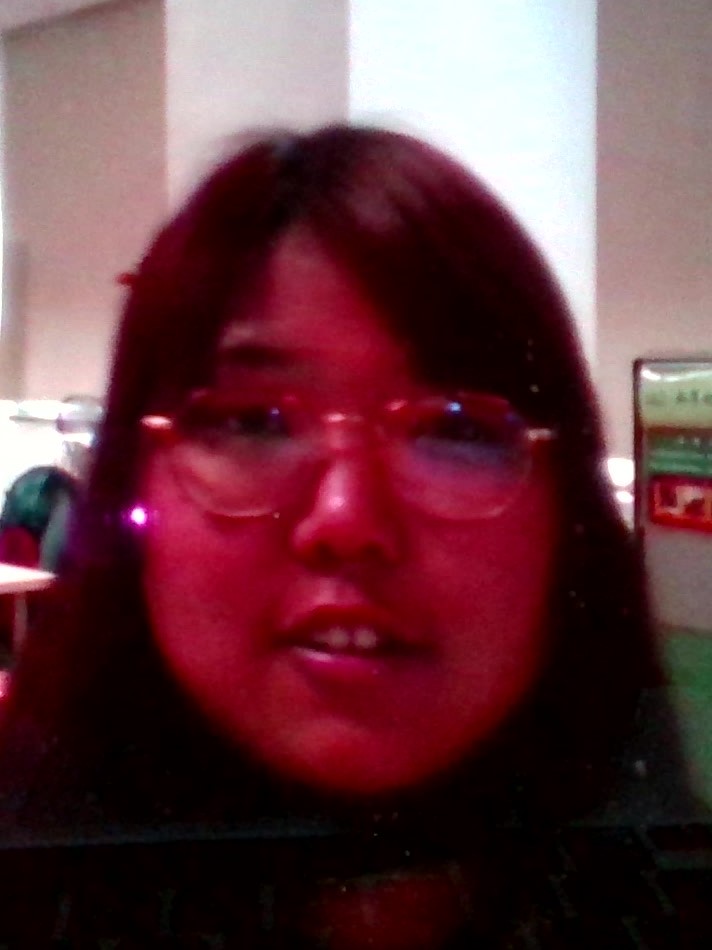}&  \includegraphics[width=0.12\textwidth,height=0.16\textwidth]{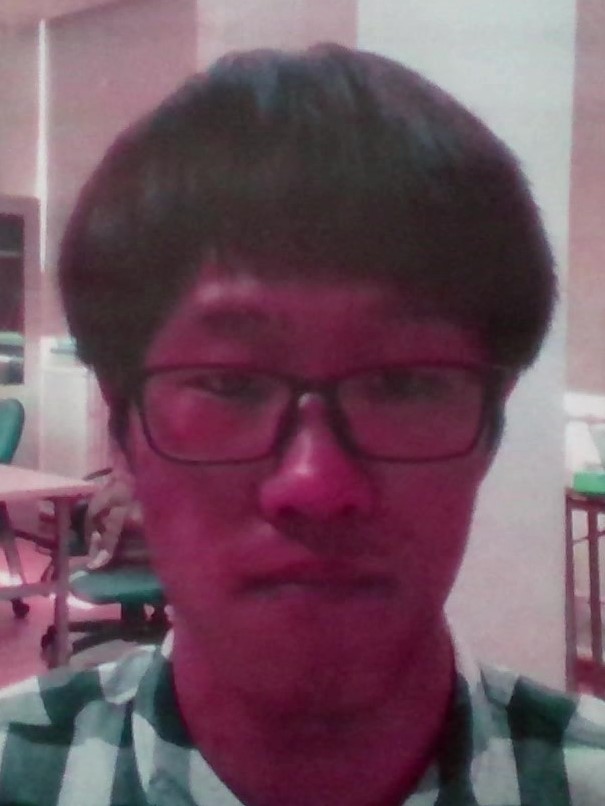}\\
\end{tabular}
\caption{Sample frames of different sessions in the proposed FA3D dataset.}
\label{fig:database}
\end{center}
\vspace{-0.7cm}
\end{figure*}

\textbf{Spectral filtering of hypergraph signals.} The convolution operator on a graph $ \ast_{\mathcal{G}} $ is first defined in the spectral domain \cite{bruna2013spectral}, specifically in the GFT domain. GFT is computed from the graph Laplacian matrix. As the graph Laplacian is symmetric and positive semi-definite, it admits a complete set of orthonormal eigenvectors. The GFT basis $\mathbf{U}$ is then the eigenvector set of the Laplacian matrix. The GFT of a graph signal $ \mathbf{x} $ is thus defined as $ \hat{\mathbf{x}} = {\mathbf{U} ^T}\mathbf{x} $, and the inverse GFT follows as $ \mathbf{x}' = \mathbf{U} \hat{\mathbf{x}} $. As an extension to hypergraphs, the normalized hypergraph Laplacian $\mathcal{L}$ defined in Eq.~\ref{eq:hyperL} is also symmetric and positive semi-definite. Thus, we extend GFT to Hypergraph Fourier Transform (HGFT), with the eigenvectors $\tilde{\mathbf{U}}$ of the normalized hypergraph Laplacian $\mathcal{L}$ as the basis.     
 
Hence, the convolution between two hypergraph signals $ \mathbf{x} $ and $ \mathbf{y} $ can be defined as the multiplication of the corresponding HGFT coefficients, followed by the inverse HGFT, i.e., 
\begin{equation}
\mathbf{x} \ast_{\mathcal{G}} \mathbf{y} 
= \tilde{\mathbf{U}}(\tilde{\mathbf{U}}^T\mathbf{x}) \odot (\tilde{\mathbf{U}}^T\mathbf{y}),
\end{equation}
where $ \odot $ is the element-wise Hadamard product. Then the spectral filtering of a graph signal $ \mathbf{x} $ by $ g_\theta $ is 
\begin{equation}
\mathbf{y} = g_{\theta}(\mathcal{L})\mathbf{x}
           = g_{\theta}(\tilde{\mathbf{U}}\mathbf{\Lambda}\tilde{\mathbf{U}}^T)\mathbf{x}
           = \tilde{\mathbf{U}}g_{\theta}(\mathbf{\Lambda})\tilde{\mathbf{U}}^T\mathbf{x}, 
\end{equation}  
where $\mathbf{\Lambda}$ is a diagonal matrix of the eigenvalues of $\mathcal{L}$.

\textbf{Chebyshev approximation for fast filtering.} The spectral filtering, however, has high computational complexity of $ \mathcal{O}(n^3) $ due to the eigen-decomposition of the Laplacian. Inspired by the truncated Chebyshev polynomials for the approximation of the spectral filtering \cite{defferrard2016convolutional}, we approximate the aforementioned hypergraph convolution by $K$-hop localized Chebyshev polynomial filtering:
\begin{equation}
\mathbf{y} = g_\theta(\mathcal{L}) \mathbf{x} 
          \approx \sum_{k=0}^{K-1}\theta_k T_k(\mathcal{L}) \mathbf{x},
\label{eq:chebyshev}
\end{equation}
where $\theta_k$ denotes the $ k $-th Chebyshev coefficient. $T_k(\mathcal{L})$ is the Chebyshev polynomial of order $ k $. It is recurrently calculated by $ T_k(\mathcal{L}) = 2\mathcal{L}T_{k-1}(\mathcal{L}) - T_{k-2}(\mathcal{L})$, where $T_0(\mathcal{L}) = 1, T_1(\mathcal{L}) = \mathcal{L}$. The computational complexity is then reduced to $ \mathcal{O}(K | \mathcal{E} | ) $. Practically, we set $K$ to 1 in order to reduce complexity.

\subsection{Feature learning}
Inspired by CNN, we apply the fully-connected layer with ReLU activation function to previous features and extract high-dimensional features, which is formulated as follows.
\begin{equation}
    \mathbf{y} = \text{ReLU}(g_\theta(\mathcal{L})\mathbf{x}\mathbf{W} + \mathbf{b}),
    \label{eq:featureLearning}
\end{equation}
where $\mathbf{W} \in \mathbb{R}^{F_1 \times F_2}$ is a matrix of learnable weight parameters, and $ F_1 $ and $ F_2 $ are the dimensions of generated features in two connected layers respectively. $\mathbf{b} \in \mathbb{R}^{n \times F_2}$ is the bias.

\begin{figure*}[t]
	\centering
	\includegraphics[width=\textwidth]{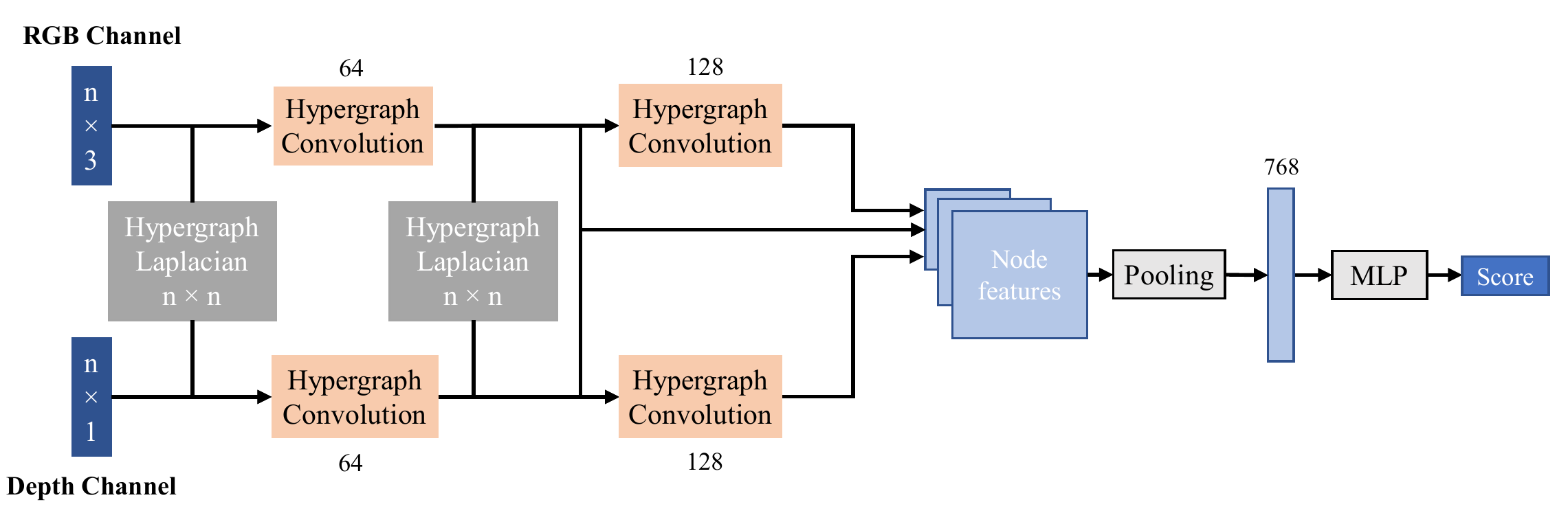}
	\caption{The details of the proposed HGCNN structure.}
	\label{fig:pipeline2}
	\vspace{-0.2cm}
\end{figure*}

In particular, as illustrated in Fig.~\ref{fig:pipeline2}, color and depth channels are fed into two similar branches without weight sharing, each composed of two hypergraph convolution layers of $(64,128)$ hidden nodes. Limited by the database scale, we only add two layers to avoid over-fitting. After that, we concatenate features from different layers as the final output of hypergraph convolution. The bypass connection enlarges the receptive field of the network. Finally, the output features are sent to the average pooling layer, which are pooled into a $768$-dimension vector. The final score is calculated by three fully-connected layers with ReLU activation function, which are composed of $(256,64,2)$ hidden nodes. We choose the cross-entropy cost function as the object function to minimize. 


\section{Dataset Details}

In order to overcome the deficiency of 3D face data, we collect a 3D face anti-spoofing database, referred to as FA3D, with color, depth and IR information. Apart from rich facial information, there exhibits high variation in our database like translation, expression and rotation. The dataset will be made public soon.  

The database contains 285 videos of 19 subjects recorded by Intel RealSense SR300. The videos include RGB videos of resolution $1920 \times 1080$, the corresponding aligned depth videos and IR videos of the same resolution\footnote{The resolution of the captured depth and IR videos is $640 \times 480$, which are then processed by the SDK of RealSense to reach the same resolution as RGB videos.}. For each subject, we collect five sections, each corresponding to a different posture, as demonstrated in Fig.~\ref{fig:database}. In section 1, the subject blinks several times; in section 2 the subject moves horizontally and vertically; in section 3 the subject moves back and forth; and in section 4 people are asked to make expressions such as smile. In section 5 the subject yaws within -15$^\circ$ to 15$^\circ$. The attacks include all the aforementioned categories we aim to address: print attacks, replay attacks and 3D mask attacks. Print attacks are based on high-resolution photos printed by Canon LBP7100, and the videos are recorded under the same settings as real ones except that we skip the rotation part. Replay attacks originate from real video clips, replayed by Macbook Pro under the same environment. For mask attacks, we employ a unique 3D latex mask and let different people wear the mask and record videos. To summarize, the attacks consist of 3-subject print attacks, 5-subject replay attacks and 4-subject mask attacks. Besides, the database contains real samples of 7 subjects. 

\section{Experiments}
We first perform detailed ablation study to examine the contributions of the proposed model components to the face anti-spoofing performance. Then we compare our results on 3D datasets and 2D datasets with other state-of-the-art approaches for comprehensive evaluation, as well as cross test. Finally, we provide some visualization and analysis of our model. 

\subsection{Experimental Setup}
\textbf{Databases} We evaluate the proposed framework on multiple attack databases for generalizability, including 3DMAD \cite{ERDOGMUS_BTAS-2013}, the proposed FA3D and Oulu-NPU \cite{boulkenafet2017oulu}. 3DMAD is the only existing publicly available 3D spoofing database, containing 17 subjects and 255 video clips with 3D masks from Thatsmyface.com. Each video is recorded by Kinect with resolution $640 \times 480$. The large-scale Oulu-NPU database consists of 5940 real and attack videos recorded with six types of smartphones under three different illumination conditions. The attack types are print and video-replay, using two printers and two display devices. The number of subjects is 55, which are divided into subject-disjoint training, development and testing. 3DMAD and our proposed FA3D are primary databases on which we conduct experiments with our 3D model. Besides, we test on the 2D dataset Oulu-NPU in order to compare with traditional 2D anti-spoofing methods for comprehensive studies. For each database, we follow the training-validation-test protocol as described. 

\textbf{Model details} The detail of our framework is depicted in Fig.~\ref{fig:pipeline2}. For each database, we prepare landmarks and hypergraphs before feeding into the network. The input then consists of 318 points with RGB and depth channels as well as the constructed hypergraph Laplacian. Specially, batch normalization is added before the activation function. During the training process, we set the learning rate according to the database, ranging from $10^{-3}$ to $10^{-4}$. Adam optimizer and Xavier initializer are utilized with batch size of $50$. 

\textbf{Evaluation metrics} Many metrics have been proposed to evaluate the performance of face anti-spoofing. In our experiments, we deploy the following commonly used metrics \cite{RN41}: Accuracy (ACC), Half Total Error Rate (HTER), Equal Error Rate (EER), Attack Presentation Classification Error Rate (APCER), Bona Fide Presentation Classification Error Rate (BPCER), Average Classification Error Rate (ACER), False Discovery Rate (FDR), True Discovery Rate (TDR) and Area Under the ROC Curve (AUC).

\subsection{Experiment Results}
\subsubsection{Ablation Study}
\begin{table}[t]
\begin{center}
\caption{Ablation study results from different models on 3DMAD}
\label{table:ablation}
\begin{tabular}{lllll}
\toprule
                        & \multicolumn{4}{c}{TDR}                                             \\ \hline
\multicolumn{1}{c}{FDR} & \multicolumn{1}{c}{1\%} & \multicolumn{1}{c}{5\%} & 10\%   & 20\%   \\ \hline
Model 1                 & 54.8\%                    & 58.4\%                   & 69.9\% & 88.4\% \\ \hline
Model 2                 & 75.7\%                  & 80.5\%                  & 84.6\% & 93.3\% \\ \hline
Model 3                 & 86.3\%                  & 93.6\%                  & 97.8\% & 98.2\% \\ \hline
Model 4                 & \textbf{97.8\%}                  & \textbf{100\%}                   & \textbf{100\%}  & \textbf{100\%}  \\ \bottomrule
\end{tabular}
\end{center}
\vspace{-0.5cm}
\end{table}


In order to research advantages of different modules of the proposed HGCNN comprehensively, we design the following incomplete models. Model 1 is the model without hypergraphs, which is equivalent to the original model except the number of vertices in each hyperedge $k$ = 0. Model 2 is our model with the depth channel removed, i.e., only the RGB cue is taken as the input. In Model 3, we replace the hypergraph representation with simple complete graphs, where the weight of each edge is assigned as the exponential function of the Euclidean distance between the two connected vertices $i$ and $j$, i.e., $w_{i,j}=\text{exp}\{-\|\mathbf{c}_i-\mathbf{c}_j\|_2^2\}$, with $\mathbf{c}_i$ and $\mathbf{c}_j$ denoting the coordinates of $i$ and $j$ respectively. The hypergraph Laplacian is then replaced with the simple graph Laplacian accordingly as used in \cite{kipf2016semi}. Model 4 is the proposed complete model. 

We test these models on the 3DMAD database and follow the protocol in \cite{ERDOGMUS_BTAS-2013}. More specifically, since the documentation doesn't define the index of  validation, we follow the leave-one-out-cross-validation (LOOCV) settings and calculate the average value. The results are reported in Tab. \ref{table:ablation}, where we calculate TDR at different FDR, the higher the better. From the results we observe that Model 1 has the worst performance, which means the hypergraph representation and convolution plays a vital role in our model. The degradation of performance in Model 2 is less significant but still large without depth, thus indicating the importance of incorporating the depth cue. When hypergraphs are replaced with simple graphs as in Model 3, the performance is better but not as good as Model 4, which validates that the hypergraph representation is superior to simple graph representation for describing higher-order relationships. Model 4 achieves the best performance of $100\%$ in TDR when FDR is beyond $5\%$.

\subsubsection{Results on 3D datasets}

\begin{table}[t]
\begin{center}
\caption{Intra-test results of different methods on 3DMAD}
\label{table:3DMAD}
\begin{tabular}{@{}llc@{}}
\toprule
\multicolumn{1}{c}{Method} & \multicolumn{1}{c}{HTER} \\ \midrule
Siddiqui et al. (HOOF) \cite{siddiqui16multifeature}              &          2.35\%             \\ 
Chingovska et al. (LBP + SVM) \cite{RN37}        & 0.95\% \\
Feng et al. (OFM) \cite{feng16jvcir}			  &         4\%                       \\
Edmunds et al. (Motion) \cite{RN28}        &        3.53\%        		 \\
Liu et al. (rPPG) \cite{liu2018remote} 				&   4.22\% 			\\
Menotti et al. (CNN) \cite{menotti2015deep} 				&   0.70\% 		\\
\hline
HGCNN                         &           \textbf{0\%}   \\     
\bottomrule
\end{tabular}
\end{center}
\vspace{-0.6cm}
\end{table}

\textbf{3DMAD} We conduct intra-test on 3DMAD evaluated by HTER, with results reported in Tab. \ref{table:3DMAD}. We follow the test protocol in \cite{ERDOGMUS_BTAS-2013}, i.e., selecting 8 subjects for training, 5 for validation, and 5 for test. From the results we observe that our model achieves the state-of-the-art performance compared with model-driven \cite{siddiqui16multifeature,RN37,feng16jvcir,RN28,liu2018remote} and CNN-based methods \cite{menotti2015deep}. The model-driven methods exploit motion (optical flow map (OFM), rPPG), texture (LBP) or multi-cue integration (HOOF), while the CNN-based one deploys the VGG-16 model and fine-tuning.

\begin{table}[t]
\begin{center}
\caption{Comparison with 2D methods on FA3D}
\label{table:FA3D_Comparison}
\begin{tabular}{@{}lcccc@{}}
\toprule
Method & APCER  & BPCER   & ACER   & ACC  \\ \midrule
LBP+SVM \cite{ERDOGMUS_BTAS-2013}    & 10.1\% & 65.4\% & 38.1\% & 66.5\% \\
FASNet \cite{Lucena2017} & 0.7\%  & 9.2\%   & 4.5\%  & 97.6\% \\
HGCNN   & \textbf{0.1\%}  & \textbf{1.6\%}   & \textbf{0.7\%}  & \textbf{99.6\%} \\ \bottomrule
\end{tabular}
\end{center}
\vspace{-1cm}
\end{table}

\textbf{FA3D} In order to evaluate more comprehensive performance on our dataset, we design different protocols for test. Protocol 1 is designed to test the generality in terms of unseen subjects, which means subjects appearing in the training data are absent in the test process and vice versa. Specifically, we randomly select 10 subjects as training data and 7 subjects for test. In protocol 2, we test the robustness to different postures. We extract one posture from each subject in the training stage, and use the other postures for test. In protocol 3, we exemplify the efficiency of depth data by splitting attack types. Mask attacks are utilized during training, while print and replay attacks are used for test. The results are reported in Tab. \ref{table:FA3D}. We can see that the accuracy is high for all the protocols. In particular, with the depth auxiliary our model only misclassifies 46 out of 12000 true frames, i.e., resulting in BPCER of $0.4\%$.

Furthermore, in order to compare our method with CNN-based methods on 3D datasets, we implement the state-of-the-art CNN-based method FASNet \cite{Lucena2017} and texture-based method LBP+SVM \cite{RN37} with RGB-only input, and conduct experiments on FA3D of protocol 1. As reported in Tab. ~\ref{table:FA3D_Comparison}, our method achieves the best performance on any metric. Also, note that BPCER is higher compared with APCER, which is due to the data skew---fake samples are 4 times more than real ones, i.e., it is aimed to minimize false acceptance rate.

\begin{table*}[t]
\begin{center}
\caption{Our intra-test results with different protocols on FA3D}
\label{table:FA3D}
\begin{tabular}{@{}cccccccc@{}}
\toprule
Protocol & Variation    & HTER  & APCER & BPCER & ACER & ACC  \\ \midrule
1        & Subjects     & 0.5\% & 0.1\% & 1.6\% & 0.7\% & 99.6\% \\
2        & Posture      & 0.6\% & 0.0\% & 1.8\% & 0.9\% & 99.3\%\\
3        & Attack Types & 0.3\% & 0.0\% & 0.4\% & 0.2\% & 99.9\%\\ \bottomrule
\end{tabular}
\end{center}
\end{table*}

\subsubsection{Cross Test on 3DMAD and FA3D}
\begin{table}[t]
\begin{center}
\caption{Cross-test results on FA3D and 3DMAD.}
\label{table:cross}
\begin{tabular}{@{}lcc@{}}
\toprule
Train-Test & FA3D-3DMAD & 3DMAD-FA3D \\ \midrule
HTER       & \textbf{34.0\%}    & 48.1\%    \\
EER        & \textbf{43.0\%}    & 49.2\%    \\
AUC        & \textbf{62.5\%}    & 51.5\%    \\ \bottomrule
\end{tabular}
\end{center}
\vspace{-0.5cm}
\end{table}
Generalizability is a primary problem in real situations. Since the illumination and depth variance is quite different, cross test is much more challenging than intra-test. we test the generality across different situations on 3DMAD and FA3D datasets, after filtering replay and print attacks in FA3D so that only mask attacks remain in both datasets. The results are reported in Tab.~\ref{table:cross}. We achieve $34.0\%$ in HTER on FA3D-3DMAD, while the inverse one only achieves $49\%$ in HTER. This is probably due to the little variance in the 3DMAD dataset, in which the illumination and postures are almost identical. 

\subsubsection{Results on 2D datasets}
To make comparison with traditional 2D-based methods, we further test our model on the widely used RGB-only dataset Oulu-NPU with Protocol 1. Note that, our method is able to achieve accuracy of $100\%$ with the depth auxiliary on 2D attacks by examining whether the input image/video has face-like depth.
For fair comparison, we remove the module of depth convolution in our model, i.e., only the RGB cue is leveraged in our method in this comparison. We compare with Baseline \cite{boulkenafet2017competition}, MixedFASNet \cite{Lucena2017} and GRADIANT\_extra \cite{boulkenafet2017competition}, and list the results in Tab.~\ref{table:oulu}. We see that HGCNN (RGB-only) achieves competitive performance with the state-of-the-art methods, with GRADIANT\_extra reaching even lower ACER. This is because this method fuses color, texture and motion information as well as exploiting both HSV and YCbCr color spaces, whereas our RGB-only HGCNN only exploits the RGB cue in a static image without temporal information.

\begin{figure}[t]
    \centering
    \subfigure[Print]{\includegraphics[width=0.13\textwidth]{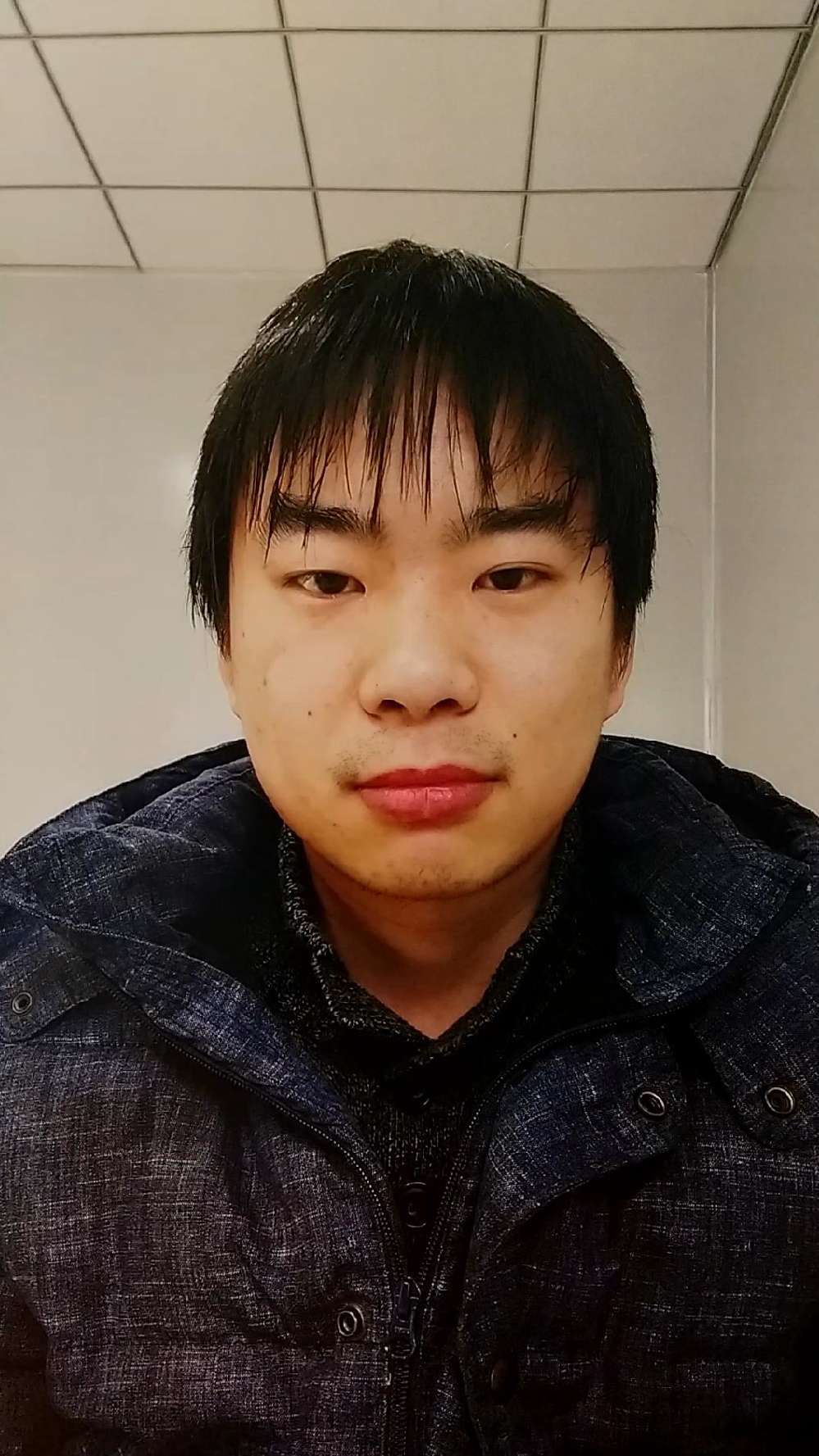}}\hspace{0.2cm}
    \subfigure[Replay]{\includegraphics[width=0.13\textwidth]{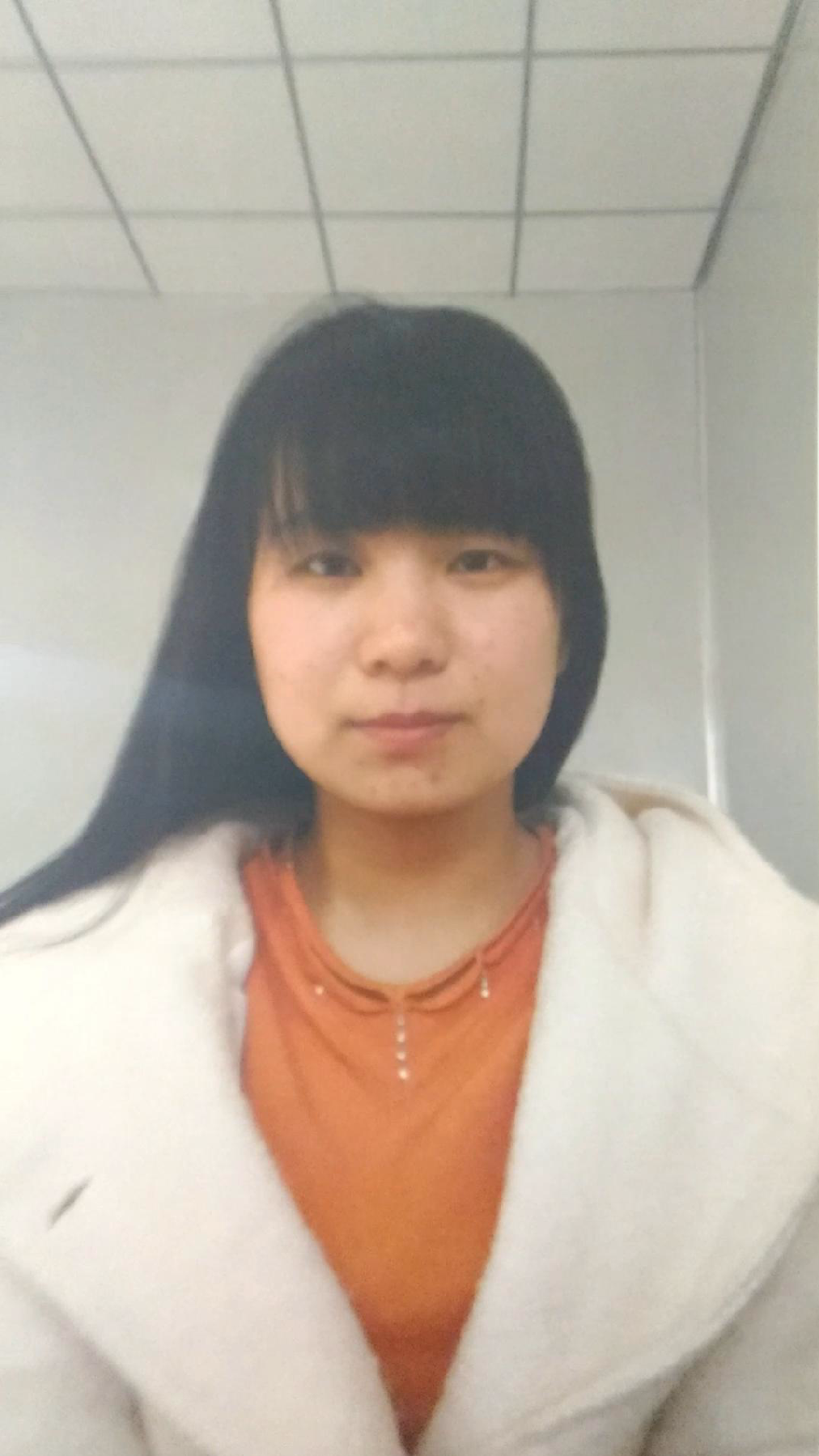}}\hspace{0.2cm}
    \subfigure[Real]{\includegraphics[width=0.13\textwidth]{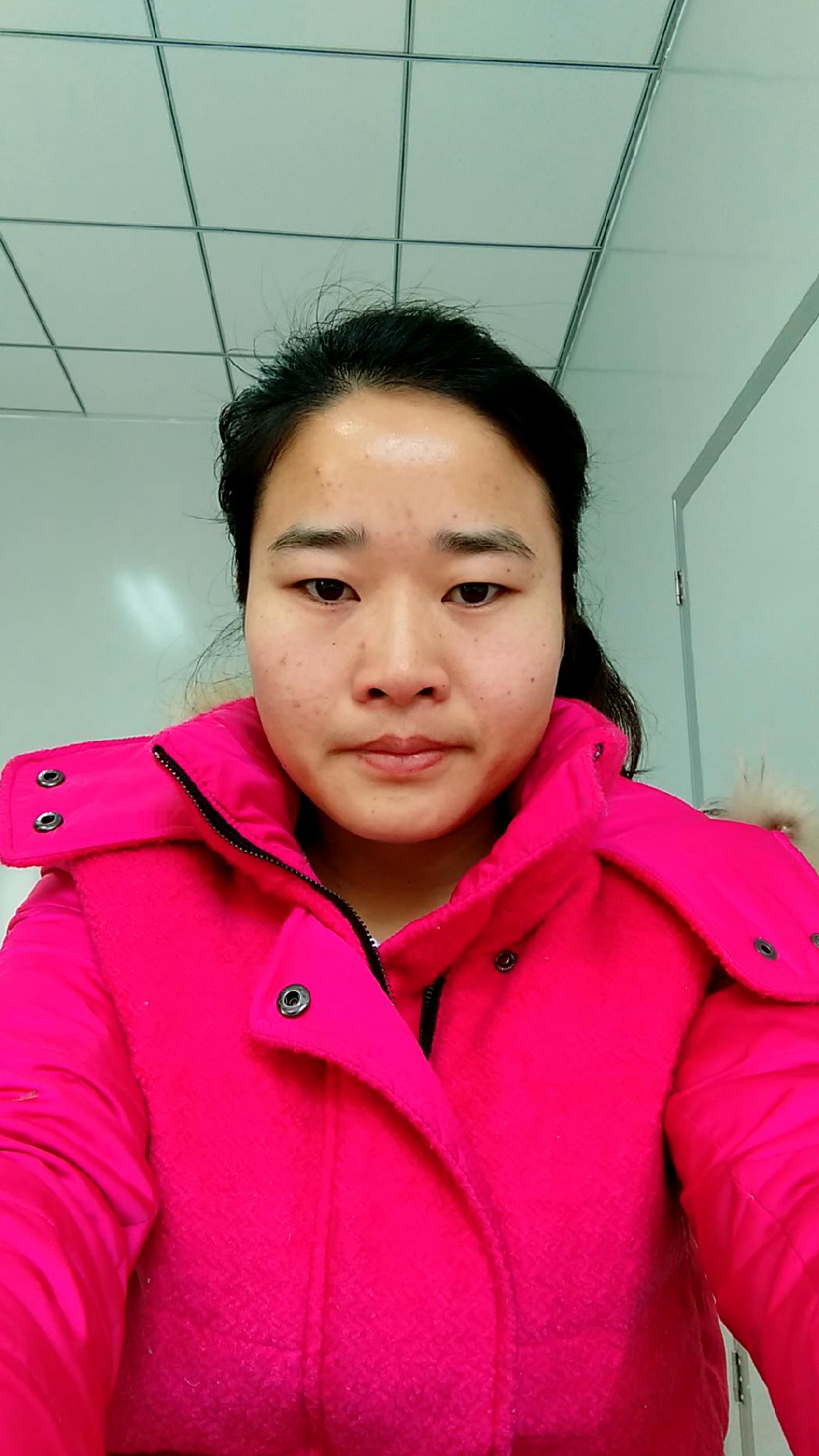}}
	\caption{Examples of failure anti-spoofing on Oulu-NPU.}
	\vspace{-0.5cm}
	\label{fig:example}
\end{figure}

As HSV aligns with human vision perception better, we further exploit HSV and RGB color spaces together, thus resulting in 6-channel images as the input. Also, we increase the degree of all the hyperedges so as to mine the textural features globally. This is referred to as HGCNN (RGB+HSV) in Tab.~\ref{table:oulu}, which outperforms all other methods. This further demonstrates the ability of extracting texture features based on color cues. Some false accepted samples are presented in Fig.~\ref{fig:example}, due to the deficiency of subtle feature extraction. For example, the replay attack in Fig.~\ref{fig:example}(b) is of low resolution, but such image quality feature is not fully exploited due to sparse landmarks.


\begin{table}[t]
\begin{center}
\caption{Results from different methods on Oulu-NPU.}
\label{table:oulu}
\begin{tabular}{@{}l|cc|c@{}}
\toprule
\multicolumn{1}{c|}{Methods} & \multicolumn{1}{c}{APCER} & \multicolumn{1}{c}{BPCER} & \multicolumn{1}{|c}{ACER} \\ \midrule
MixedFASNet \cite{Lucena2017}                         &       0.0\%        & 17.5\%    & 8.8\%      \\ 
GRADIANT\_extra \cite{boulkenafet2017competition}                  &          7.1\%                 &          5.8\%      & 6.5\%  \\ 
Baseline \cite{boulkenafet2017competition}  &          5.0\%                &           20.8\%       &12.9\% \\ \midrule
HGCNN (RGB)                  &          6.5\%                 &          10.1\%      & 8.3\%  \\ 
HGCNN (RGB+HSV)                  &          6.0\%                &           6.8\%       &\textbf{6.4}\% \\
\bottomrule

\end{tabular}
\end{center}
\vspace{-0.5cm}
\end{table}

\subsubsection{Visualization and Analysis}
\begin{figure}[t]
    \centering
    \subfigure[Layer 0 (Input)]{\includegraphics[width=0.15\textwidth]{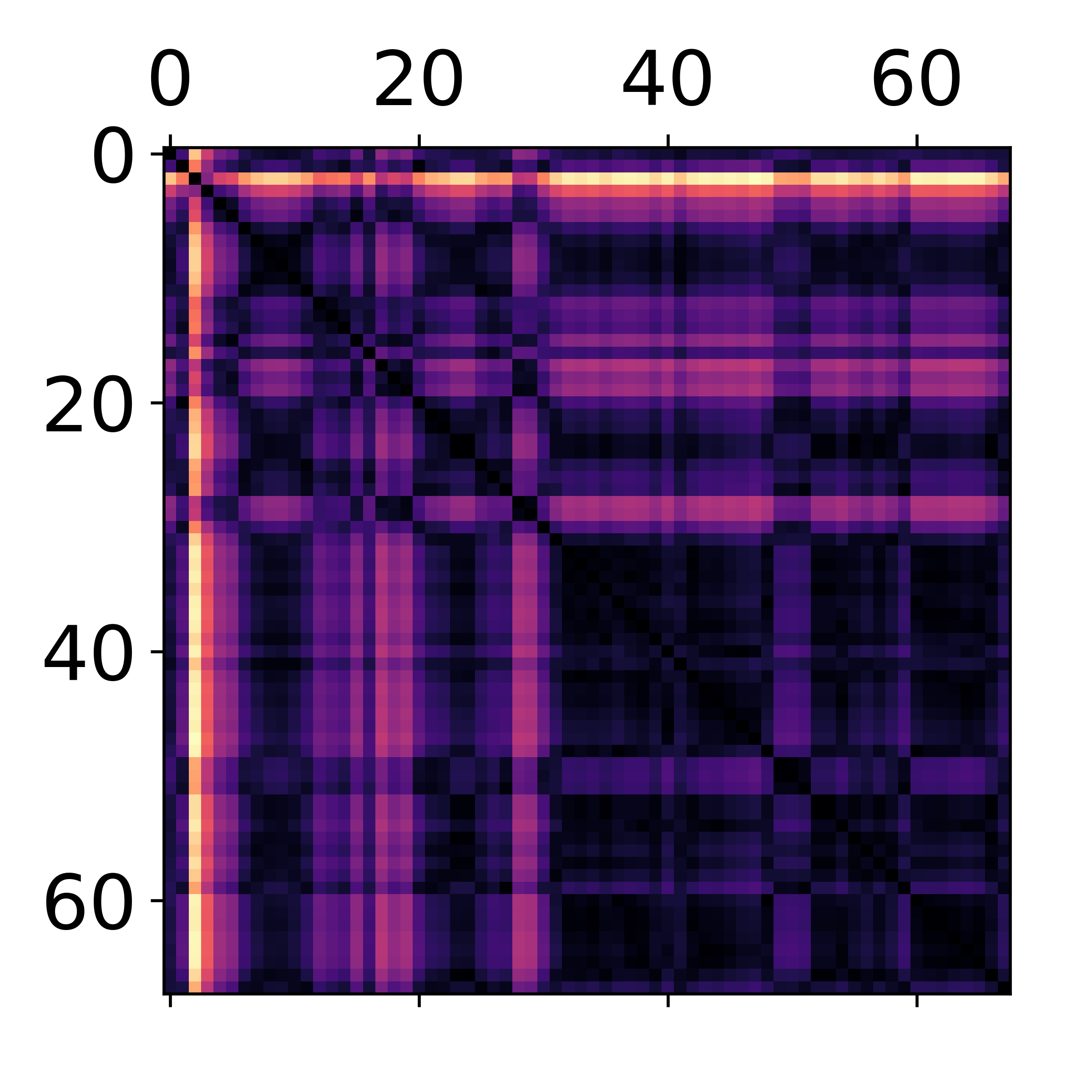}}
	\subfigure[Layer 1]{\includegraphics[width=0.15\textwidth]{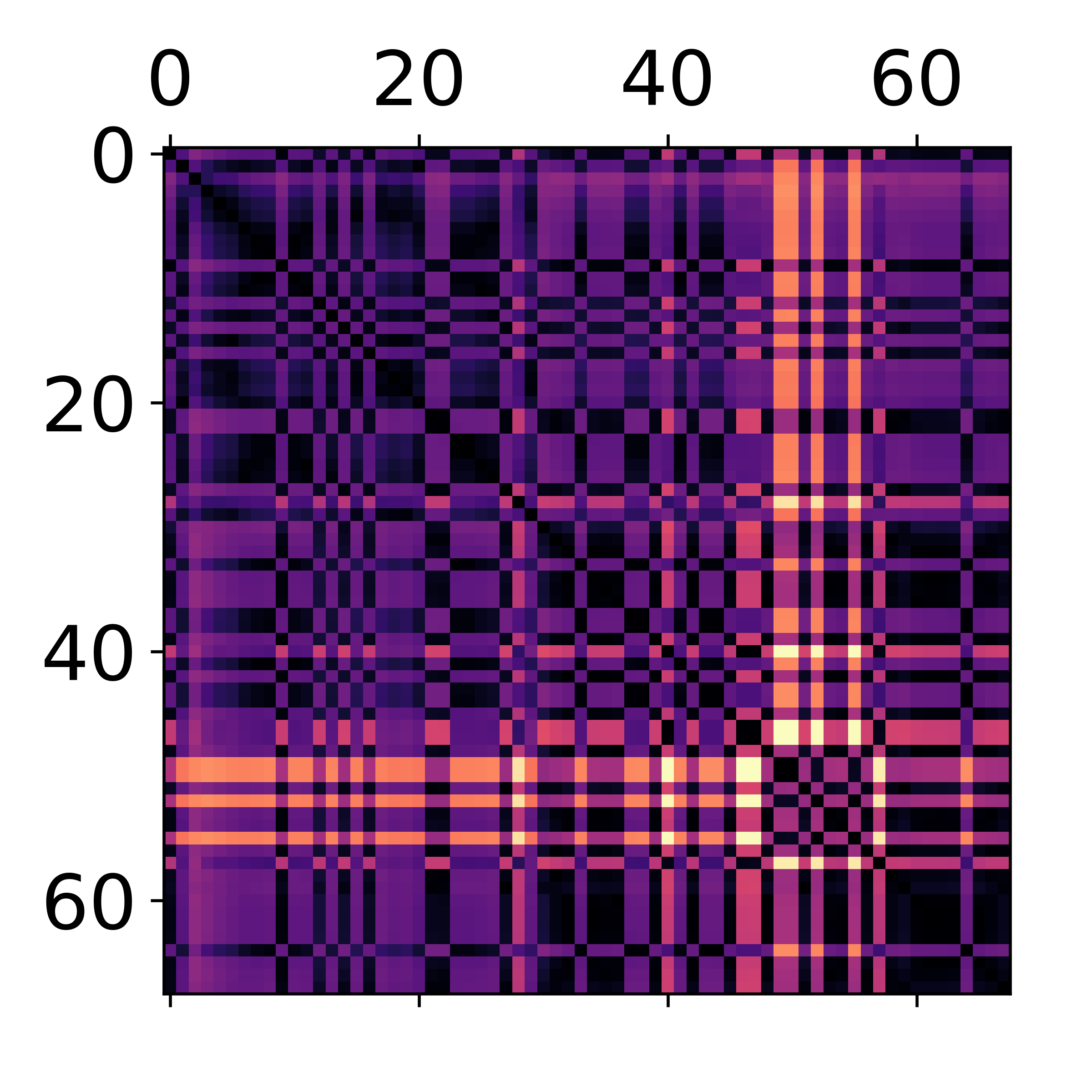}} 
    \subfigure[Layer 2]{\includegraphics[width=0.15\textwidth]{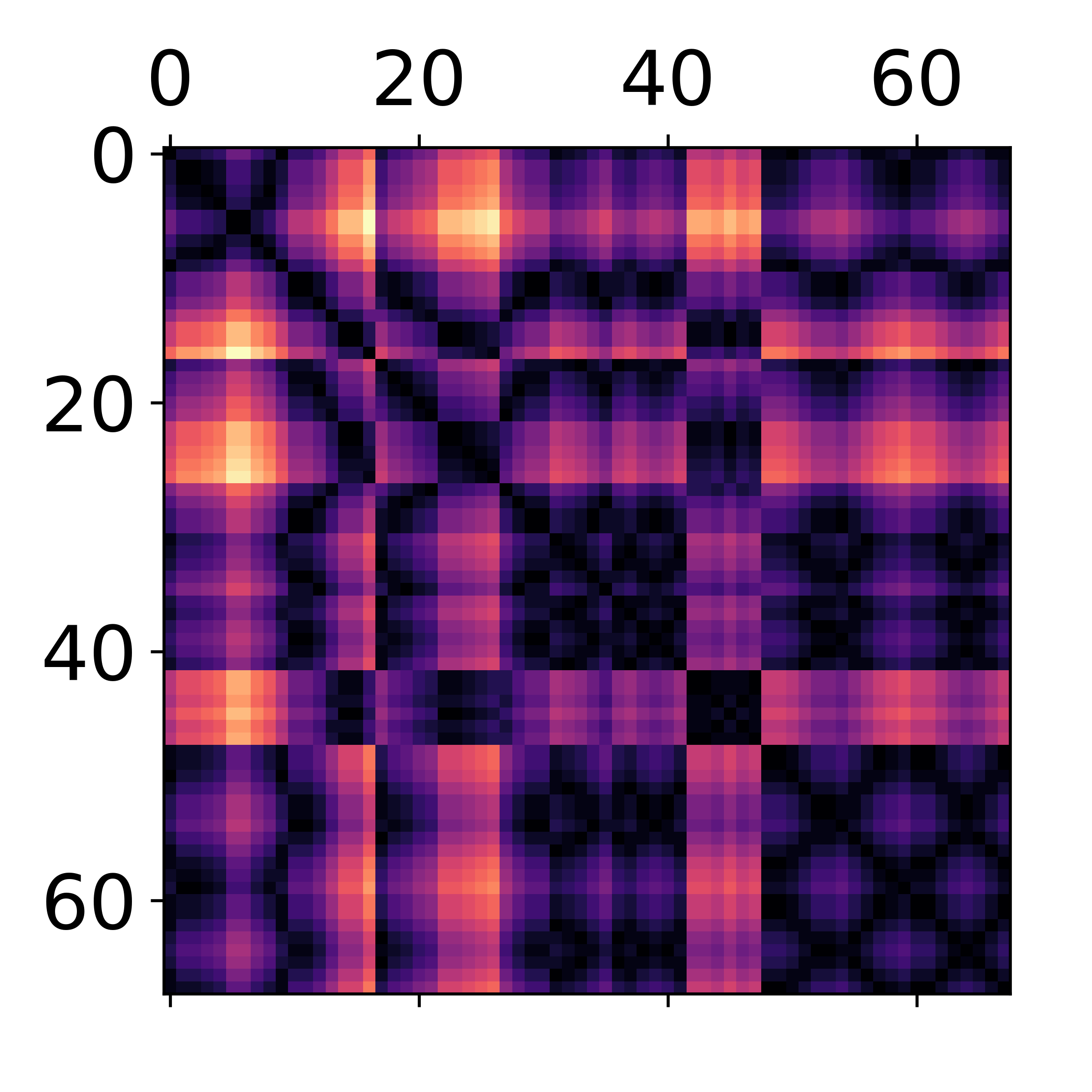}}
	\caption{Euclidean distance among 68 landmarks. Darker colors denote smaller distance.}
	\label{fig:visual}
	\vspace{-0.5cm}
\end{figure}

To interpret the graph structure more vividly, we visualize the latent feature space of different layers in Fig.~\ref{fig:visual}, where darker colors represent smaller distance. We observe that point features in the input are not quite distinguished from each other, but after hypergraph convolution points tend to keep similar features with adjacent ones, especially in certain regions like the mouth, nose and eyes. At Layer 2, points within same facial features resemble to each other, which is reflected in Fig.~\ref{fig:visual} that the color of certain regions is very dark, such as 0-17 (the edge of left face) and 48-68 (the edge of mouth). 

\section{Conclusion}
We propose hypergraph-based convolutional neural networks for 3D face anti-spoofing, which provides computation-efficient and posture-invariant face representation and enables learning of high-order relationship among samples via hypergraph convolution. Also, we exploit fusion of RGB and depth cues by performing the same hypergraph convolution on them, and identify the importance of the depth auxiliary for 3D face anti-spoofing. Besides, we collect a 3D face attack database that contains more subjects and variations than prior 3D face attack databases. Extensive experiments demonstrate the superiority of our method.

{\small
\bibliographystyle{ieee}
\bibliography{egbib}
}
\end{document}